\documentclass{article}
\usepackage{amssymb}
\usepackage{amsmath}
\usepackage{graphicx}
\usepackage{wrapfig}

\usepackage{xcolor}

\definecolor{temp_green}{HTML}{2E7D32}   
\definecolor{temp_red}{HTML}{C62828}     
\definecolor{temp_yellow}{HTML}{F9A825}  
\definecolor{blue}{HTML}{1565C0}         
\definecolor{purple}{HTML}{6A1B9A}
\definecolor{orange}{HTML}{D84315}
\definecolor{teal}{HTML}{00796B}
\definecolor{pink}{HTML}{AD1457}
\definecolor{red_best}{HTML}{C60000}
\usepackage[preprint]{corl_2025} 
\usepackage{subcaption}
\usepackage{caption}
\usepackage{adjustbox}
\usepackage{hyperref}
\usepackage[capitalize]{cleveref}

\usepackage{multirow}
\usepackage{multicol}
\usepackage{booktabs}
\usepackage{pifont}
\usepackage{enumitem}

\crefformat{subfigure}{Fig.~\ref{#1}}
\crefname{algorithm}{Alg.}{Algs.}
\crefname{section}{Sec.}{Secs.}
\crefname{table}{Tab.}{Tabs.}
\crefname{figure}{Fig.}{Fig.}
\crefname{appendix}{Appx.}{Appx.}

\newcommand{\qpos}[0]{\mathbf{x}}
\newcommand{\qvel}[0]{\dot{\mathbf{x}}}
\newcommand{\qacc}[0]{\ddot{\mathbf{x}}}
\newcommand{\jpos}[0]{\mathbf{q}}
\newcommand{\jvel}[0]{\dot{\mathbf{q}}}

\newcommand{\refpos}[0]{\qpos_{\rm ref}}
\newcommand{\refvel}[0]{\qvel_{\rm ref}}
\newcommand{\refacc}[0]{\qacc_{\rm ref}}

\newcommand{\vanilla}[0]{\textcolor{teal}{vanilla}}
\newcommand{\robust}[0]{\textcolor{orange}{robust}}
\newcommand{\builtin}[0]{\textcolor{pink}{builtin}}

\newcommand{\yes}{\textcolor{teal}{\ding{51}}}
\newcommand{\no}{\textcolor{red}{\ding{55}}}

\newcommand{\method}{FACET}

\title{FACET: \underline{F}orce-\underline{A}daptive \underline{C}ontrol via Imp\underline{e}dance Reference \underline{T}racking for Legged Robots}

\author{Botian Xu$^{*1}$ \quad
Haoyang Weng$^{*1}$ \quad
Qingzhou Lu$^{*1}$ \thanks{equal contribution.} \quad 
Yang Gao$^{123}$ \quad
Huazhe Xu$^{123}$ \\ 
$^1$Tsinghua University, $^2$Shanghai Qizhi Institute, $^3$Shanghai AI Lab \\
\texttt{Project Website: \url{https://facet.pages.dev/}}
\vspace{-16pt}
}

\begin{document}
\maketitle


\begin{figure}[h]
\vspace{-12pt}
    \centering
    \includegraphics[width=0.95 \linewidth]{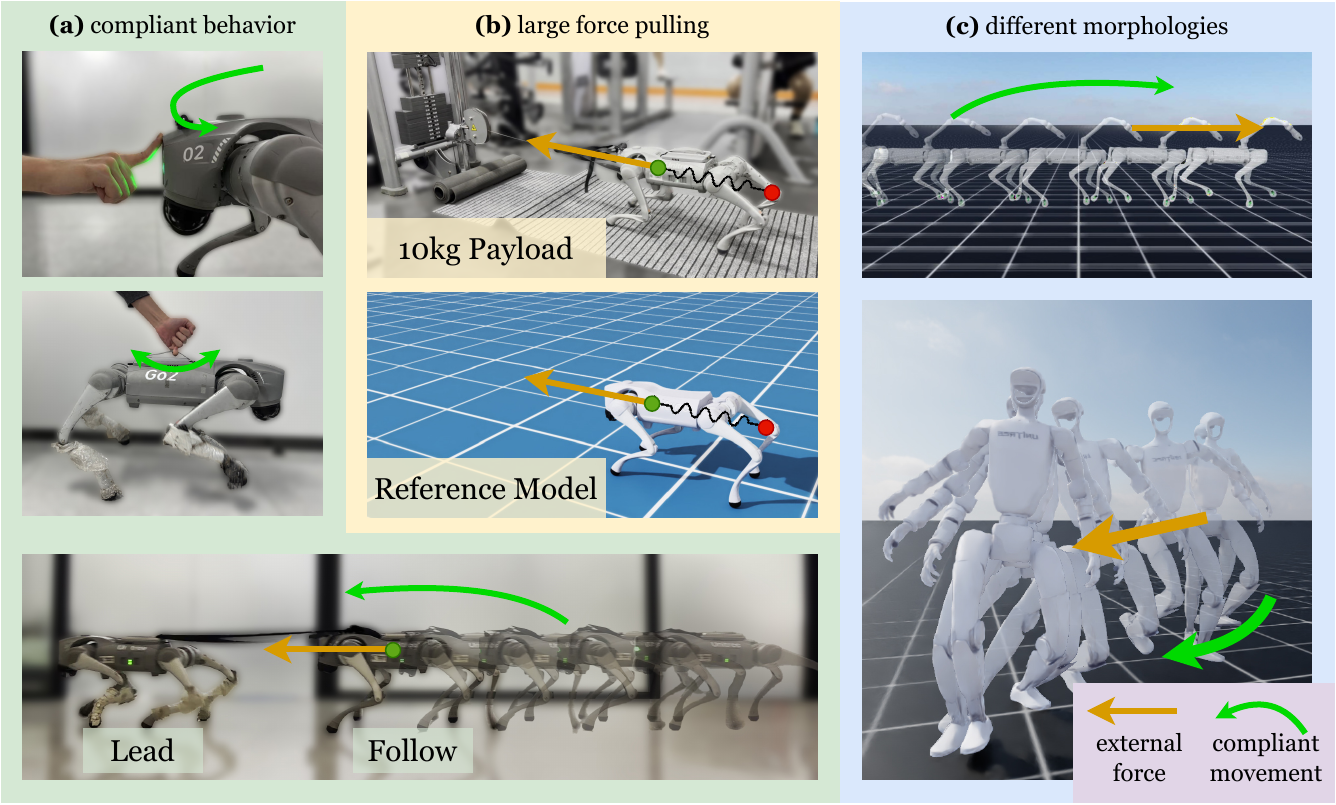}
    \caption{
    Inspired by impedance control,
    \textbf{\method{}} enables task-space variable compliance and force-adaptive control on legged robots by imitating a reference spring-mass-damper model (b) using reinforcement learning.
    A high compliance allows the robot to be stopped or kinesthetically guided with ease (a), while a high stiffness allows the robot to exert large forces when pushing/pulling a payload (b). The framework applies to different morphologies and more complex configurations (c).}
    \label{fig:head}
    \vspace{-9pt}
\end{figure}

\begin{abstract}
Reinforcement learning (RL) has made significant strides in legged robot control, enabling locomotion across diverse terrains and complex loco-manipulation capabilities.
However, the commonly used position or velocity tracking-based objectives are agnostic to forces experienced by the robot, leading to stiff and potentially dangerous behaviors and poor control during forceful interactions.
To address this limitation, we present \emph{\underline{F}orce-\underline{A}daptive \underline{C}ontrol via Imp\underline{e}dance Reference \underline{T}racking} (\method{}). Inspired by impedance control, we use RL to train a control policy to imitate a virtual mass-spring-damper system, allowing fine-grained control under external forces by manipulating the virtual spring. 
In simulation, we demonstrate that our quadruped robot achieves improved robustness to large impulses
(up to 200 Ns)
and exhibits controllable compliance, achieving an 80\% reduction in collision impulse. The policy is deployed to a physical robot to showcase both compliance and the ability to engage with large forces 
by kinesthetic control
and pulling payloads up to $2/3$ of its weight. Further extension to a legged loco-manipulator and a humanoid shows the applicability of our method to more complex settings to enable whole-body compliance control.


\end{abstract}

\keywords{Reinforcement Learning, Legged Robot Control, Impedance Control} 



\section{Introduction}

Reinforcement learning (RL) has powered many recent advancements in legged robot control, showcasing remarkable agility, robustness, and expressiveness. RL-based locomotion policies enable robots to navigate a diverse range of terrains \cite{Lee_2020,choi2023learning} and tackle complex parkour challenges \cite{zhuang2023robot, cheng2024extreme, hoeller2024anymal}. Additionally, RL-based loco-manipulation has also received increasing attention for its potential to achieve whole-body coordination to fully leverage the degrees of freedom offered by the robot. Successful applications include pick and place \cite{fu2023deep, pan2024roboduet, portela2024learning}, door-opening \cite{he2024learning, wang2024arm}, and ball shooting \cite{fey2025bridging}.

Despite the impressive progress, existing RL-based controllers often overlook or greatly simplify interactions involving external forces. Specifically, most locomotion controllers are trained to track commanded velocities regardless of the external force experienced, resulting in stiff and jerky motions \cite{hartmann2024deep}. Similarly, manipulation controllers are trained to track commanded end-effector poses, often assuming that the forces involved are negligible or can be easily counteracted, e.g., the objects to interact with have negligible mass. The limited consideration of forceful interaction poses issues when the robot operates in complex, human-populated environments. The robot's response when intervened by humans or upon collision is either stiff or unspecified, leading to potential damage to both the robot and its surroundings.

To address these limitations and provide more fine-grained control for legged robots under forceful interactions, we propose a reinforcement learning framework named \emph{\underline{F}orce-\underline{A}daptive \underline{C}ontrol via Imp\underline{e}dance Reference \underline{T}racking} (\method{}). 
Impedance control \cite{Hogan1985ImpedanceCA} has been widely used in robotic manipulation to control the end-effector with indirect force specification. Inspired by its formulation, we model the robot's behavior with a virtual mass-spring-damper system and utilize RL to train the robot to follow its dynamics. 
To achieve efficient training, we design a reference model, i.e., a virtual mass-spring-damper system defined on the center of mass (CoM) of the robot, to generate smooth tracking targets and use a teacher-student training recipe for sim-to-real transfer.

Experiments on multiple robots demonstrate the advantage of our force-adaptive controllers in terms of improved robustness and controllable compliance: a quadruped robot can survive impact forces exceeding twice its weight and significantly reduce the force required to either stop or move it. We also showcase an extension to a legged loco-manipulation system to underscore its potential for more complex forceful interaction scenarios.

The contributions of this work are summarized as follows:
\begin{itemize}[leftmargin=*]
  \item We propose \method{} to enable compliant and force-adaptive behaviors on legged robots by learning to imitate the dynamics of a virtual mass-spring-damper model. The key technical novelty of our method is to leverage a reference model to generate position–velocity tracking targets under external perturbations, while exposing the virtual impedance parameters \((\qpos_{\rm des}, K_p, K_d)\) as part of the control interface. This implicit force command interface enables intuitive simultaneous regulation of motion and interaction forces through simple manipulation of the spring–damper gains.
  \item We validate \method{} across three robot platforms (quadruped Go2, humanoid G1, and quadruped with mounted arm B1+Z1) in simulation and on real quadruped robot. In simulation, we show that \method{}: (i) survives much larger external impulses by following rather than resisting them; (ii) produces significantly smaller collision impulses against a wall by tuning stiffness; and (iii) outperforms baselines and ablations in tracking error and success rate. On hardware, we demonstrate intuitive compliant following under manual guidance and reliable pulling of payloads up to 10 kg, confirming that \method{} delivers robust, soft, and high-force interactions.  
\end{itemize}

\section{Related Work}
\paragraph{Reinforcement learning for legged robot control.}Reinforcement learning has become a promising tool for training locomotion controllers for quadrupedal and bipedal legged robots in various scenarios \cite{Lee_2020, choi2023learning}, especially favored for agility \cite{cheng2024extreme, he2024agile}. For the robot to operate reliably in complex environments, many prior works have aimed to improve robustness and adaptability under external forces by applying perturbations during training \cite{hartmann2024deep, Shi2024RethinkingRA}, enhancing state estimation \cite{Xiao2024PALOCOLP,xiao2025learning}, or learning some special behaviors to avoid damage \cite{wang2024guardians, shi2024robust}. However, as discussed in Deep Compliant Control (DMC) \cite{hartmann2024deep}, the commonly used velocity tracking reward function produces highly stiff control policies. This issue arises because the velocity tracking objective does not specify appropriate responses to external forces or always seeks to reject. Although DMC identified this issue and designed a recovery stage after the perturbation is applied, it does not offer a principled way to control the level of compliance. The same issue also happens in RL-based loco-manipulators, which use position tracking as the objective for end-effector control \cite{fu2023deep,pan2024roboduet,portela2025wholebodyendeffectorposetracking}.



\paragraph{Force and impedance control with reinforcement learning.}Hybrid position-and-force control is desirable for many robotic tasks requiring forceful interactions with the environment.
Direct force control and impedance control \cite{Hogan1985ImpedanceCA} have been well studied in robotic manipulation in both model-based and learning-based settings \cite{hou2024adaptive,martin2019variable,beltran2020learning,kalakrishnan2011learning,chen2025dexforce}, but remain undeveloped for legged robots.
The complex and contact-rich movement of the legs makes it difficult to accurately measure or track the required ground reaction force, underscoring the potential of RL-based controllers.
Some prior works explored torque-based action spaces for legged robots \cite{chen2023torque, sood2024decap, kim2023torque} to improve motion smoothness, robustness, and sim-to-real transfer. While introducing compliance to the joint space, they do not directly lead to compliance motion in Cartesian space.
Research in physics-based character control \cite{lee2022deep} introduced compliance to motion tracking controllers by altering the reference motion. Closest to our work, \cite{portela2024learning} proposed a force control objective for quadrupedal loco-manipulation systems \cite{fu2023deep, liu2024visual, pan2024roboduet, ha2024umi}. Their method allows users to command the force at the end-effector, greatly enhancing the capability for forceful interaction. However, it requires explicitly switching between a position-tracking mode and a force mode. Meanwhile, it assumes the system is near static, such that the force at the end-effector can be approximated by the external force applied.
A comparison of different objectives used in RL-based control policies is shown in \cref{tab:comparison}.

\begin{table}[h]
\centering
\small
\caption{Comparison of the control capabilities achieved by different training objectives.}
\begin{tabular}{@{}ccccccc@{}}
\toprule
Method            & Position & Velocity & Compliance & Force & Inertia$^\star$ & Dynamic \\ \midrule
Position tracking & \yes              & \no               & \no         & \no            & \no              & -                    \\
Velocity tracking, e.g. \cite{Lee_2020,choi2023learning} & \yes              & \yes              & \no         & \no            & \no              & -                    \\
Deep Compliant Control \cite{hartmann2024deep}        & \yes              & \yes              & \yes$^*$       & \no            & \no              & -                    \\
Learning Force Control \cite{portela2024learning}            & \yes              & \yes              & \yes        & \yes           & \no              & \yes$^\dagger$                  \\
\method{}              & \yes              & \yes              & \yes        & \yes           & \yes             & \yes                   \\ \bottomrule
\end{tabular}
\parbox{\textwidth}{
\small
\indent $^\star$ inertia shaping allows for manipulating a virtual mass of the system to control its acceleration behavior\\
\indent $^*$ the compliance does not have a principled form and can not be controlled after training\\
\indent $^\dagger$ a near-static assumption is needed for the force output to be approximated by external forces applied
}
\label{tab:comparison}
\vspace{-10pt}
\end{table}



\section{Background}
\paragraph{Impedance control for robotic manipulation.}
Compliance and force control are crucial in many manipulation tasks requiring complex forceful interaction. Impedance control \cite{Hogan1985ImpedanceCA} can achieve task space compliance through an indirect force control interface. The controller takes as input the desired position $\qpos_{\rm des}$ and velocity $\qvel_{\rm des}$ of a \emph{setpoint}, along with stiffness $K_p$ and damping $K_d$ that define a virtual spring-damper system. The goal is to output joint-space torques that produce an end-effector force/torque in task space according to:  
\begin{equation} f_{\rm spring} = K_p (\qpos_{\rm des} - \qpos) + K_d (\qvel_{\rm des} - \qvel)  \label{eq:impedance_force_control},
\end{equation}
where $\qpos$ and $\qvel$ denote the translational and rotational position and velocity of the body.

For fixed-base systems such as robotic arms, impedance control can be implemented by mapping the task-space force $f_\text{spring}$ into joint-space torques. 
Rather than commanding a sequence of positions directly, the user controls a virtual spring by appropriately adjusting the setpoints and the impedance gains $K_p, K_d$, then the controller will generate desired interaction forces \cite{manipulation} according to \cref{eq:impedance_force_control} to manipulate an object. 

\section{Method}

\begin{figure}[h]
    \centering
    \includegraphics[width=\linewidth]{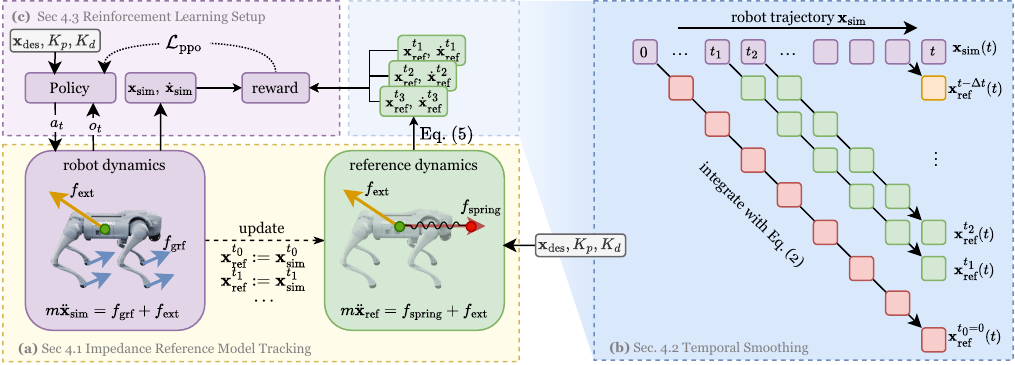}
    \caption{Method overview.
    \textbf{(a)}: The policy is trained to imitate the dynamics of an impedance reference model, i.e., a virtual mass-spring-damper system defined on the CoM of the robot (\cref{sec:ref_model_tracking}).
    \textbf{(b)}: The reference dynamics is integrated from different starts along the robot trajectory to generate the reference targets, which provides tracking rewards for the policy. (\cref{sec:temporal_smoothing}).
    \textbf{(c)} The policy receives the same set of impedance parameters $(\qpos_{\rm des}, K_p, K_d)$ and is optimized to produce the same trajectory as that of the reference model (\cref{sec:rl_setup}).}
    \label{fig:overview}
    \vspace{-10pt}
\end{figure}

\subsection{Impedance Reference Model Tracking}
\label{sec:ref_model_tracking}




Impedance control has proven effective in achieving compliant and force-adaptive behaviors for robotic manipulators. Inspired by its success, we extend impedance control principles to legged robots, aiming to regulate the center of mass (CoM) dynamics through a virtual mass-spring-damper model, so that the CoM responds as if tethered to its target by a virtual spring.

To realize this, we define a reference model whose dynamics satisfy
\begin{equation}
m \ddot{\qpos}_{\rm ref} = f_{\rm spring} + f_{\rm ext} = K_p (\qpos_{\rm des} - \qpos_{\rm ref}) + K_d (\dot{\qpos}_{\rm des} - \dot{\qpos}_{\rm ref}) + f_{\rm ext}
\label{eq:ref_model}
\end{equation}
where $m$, $K_p$, and $K_d$ are the virtual mass, stiffness, and damping; $(\qpos_{\rm des}, \qvel_{\rm des})$ the commanded setpoint; $f_{\rm ext}$ the external forces and torques applied to the system, e.g., a push or kick at the trunk, while the robot is modeled with a simplified dynamics:
\begin{equation}
    m\qacc_{\rm sim}=f_{\rm grf} + f_{\rm ext}.
\end{equation}

Our policy thus takes as input the impedance parameters $(\qpos_{\rm des}, K_p, K_d)$, in contrast to velocity-tracking controllers or direct force-control approaches, which use explicit velocity or force targets as their input interface. This indirect force commanding interface enables 
a variety of intuitive, force-centric tasks. For example, by setting $K_p = 0$, a human can easily guide the robot with minimal effort with a string attached. Additionally, kinesthetic teaching for loco-manipulators becomes feasible for wall-swiping tasks by moving the setpoint along the surface while maintains a compliant end-effector contact force.

As illustrated in \cref{fig:overview} (lower left), both the robot’s CoM and the reference model experience the same external perturbation $f_{\rm ext}$. The reference model additionally experiences the spring force $f_{\rm  spring}$ while the robot experiences ground reaction forces $f_{\rm grf}$ through contact with the ground.
In principle, realizing the dynamics in \cref{eq:ref_model} requires controlling the robot to generate $f_{\text{grf}}$ through feet contact to match the virtual spring force $f_{\text{spring}}$. However, this is challenging due to the complex contact dynamics and various constraints associated with the robot. Reliable sensing of $f_{\rm ext}$ is also infeasible due to the absence of force sensors in most robots.

To circumvent these issues, our method resorts to RL to track the reference trajectory generated by the virtual mass-spring-damper system. Specifically, we integrate the reference dynamics \cref{eq:ref_model} over time to produce smooth position and velocity trajectories $(\qpos_{\text{ref}}, \dot{\qpos}_{\text{ref}})$,
yielding the objective:
\begin{align}
 \mathcal{L}_{\text{track}}  = & \|\qpos_{\text{sim}} - \qpos_{\text{ref}}\|^2_2 \\
 + & \|\qvel_{\text{sim}} - \qvel_{\text{ref}}\|^2_2
\end{align}
This approach bypasses the need for direct force or acceleration tracking and enables stable and robust training despite the complex contact dynamics inherent to legged locomotion.

\subsection{Temporal Smoothing of Tracking Targets}
\label{sec:temporal_smoothing}
As illustrated in \cref{fig:overview} (b), we obtain the reference state at time $t$ by integrating \cref{eq:ref_model}:
\begin{align}
    \refvel^{t_0}(t)=\qvel_{\rm sim}(t_0) + \int_{t_0}^t \refacc(t) dt, ~\refpos^{t_0}(t)=\qpos_{\rm sim}(t_0) + \int_{t_0}^t \refvel^{t_0}(t) dt.
\end{align}
where $\qpos_{\rm sim}(t_0)$ and $\qvel_{\rm sim}(t_0)$ represent the robot’s actual state at timestep $t_0$.



Integrating from the beginning of each episode gives us the \textcolor{temp_red}{open-loop tracking target} $(\refpos^{t_0=0}(t),\refvel^{t_0=0}(t))$. It faithfully follows \cref{eq:ref_model} but does not consider the gap between the robot dynamics and the reference model caused by physical constraints, e.g., the acceleration a robot can produce depends on its motor capabilities and contact state with the ground. In contrast, integrating from the most recent states leads to \textcolor{temp_yellow}{closed-loop tracking target} $(\refpos^{t-\Delta t}(t),\refvel^{t-\Delta t}(t))$, which is adaptive to the robot's real state but can be noisy and short-sighted to provide sufficient reward signal for RL. Here, $\Delta t=0.02s$ is the control interval of the policy.

To address these issues, we introduce \emph{temporal smoothing}: instead of a single target, we use a mixture of targets obtained by integrating from different time steps $t'\in\{t_1,t_2,\cdots,t_M\}$ along the robot's trajectory.
The reward function is then defined to have the form:
\begin{equation}
    r_t=\frac{1}{M}\sum_{t'} \exp(-||\qpos_{\rm sim}(t)-\refpos^{t'}(t)||^2_2)+\exp(-||\qvel_{\rm sim}(t)-\refvel^{t'}(t)||^2_2).
    \label{eq:reward}
\end{equation}
In our implementation, we use a \textcolor{temp_green}{smoothed tracking targets} with $t'\in\{t-8\Delta t, t-16\Delta t,t-32\Delta t\}$. Note that choosing $t'\in\{0\}$ or $t'\in\{t-\Delta t\}$ recovers the open-loop or close-loop case, respectively.

\subsection{Reinforcement Learning for Impedance Tracking}
\label{sec:rl_setup}

\begin{wrapfigure}{r}{0.45 \textwidth}
    \centering
    \vspace{-15pt}
    \includegraphics[width=\linewidth]{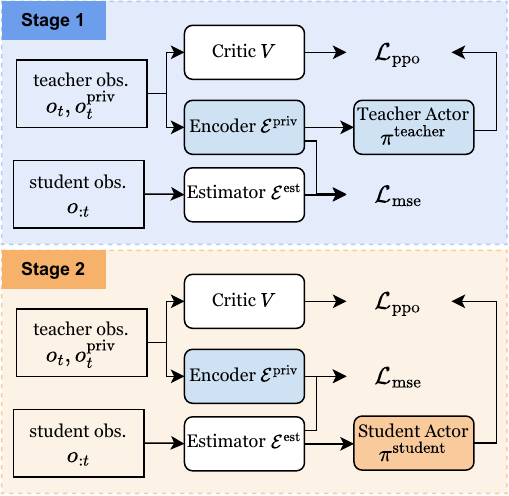}
    \small
    \caption{We train a \textbf{state estimator} $\mathcal{E}^{\text{est}}$ to predict the feature extracted by the \textbf{privileged encoder} $\mathcal{E}^{\text{priv}}$. In the second stage, the student actor $\pi^{\rm student}$ is initialized with the teacher's parameters and continues to be finetuned. }
    \vspace{-20pt}
    \label{fig:teacher-student}
\end{wrapfigure}

\paragraph{Observation and action spaces} Following \cref{eq:ref_model}, the command input to the robot is $\mathbf{u}=(\qpos_{\rm des}, K_p, K_d, m)$. For a legged robot, we are concerned with the planar translation and yaw-rotation $\qpos_{\rm des}=(x, y, \theta_z)$. The policy is trained using a teacher-student framework: the student observes the command, proprioceptive information, and previous actions $o_t=(\mathbf{u}, \jpos, \jvel, \mathbf{g}, a_{t-3:t-1})$, while the teacher additionally observes $o^{\text{priv}}_t=(\qvel, \qacc, \boldsymbol{\tau}, f_{\rm ext}, \mathbf{c})$. At each time step, the policy outputs action $\mathbf{a}_t$, which are translated to joint position targets $\jpos_{\rm des}$ and then tracked by a PD controller.
During training, we sample various command and external force distributions to expose the policy to a variety of profiles for forceful interactions. The details of sampling distributions can be found in \cref{appendix:sampling_details}.




\paragraph{Sim-to-real transfer via teacher-student training}


In the real-world, direct measurements of body velocity and external forces are unavailable due to the lack of such sensors. The reference model is therefore also unavailable. During training, a control policy must learn to infer such \emph{privileged information} to be deployed in the real world.

We adopt a two-stage teacher-student framework similar to RMA \cite{kumar2021rma}. When training the teacher policy $\pi^{\rm teacher}$ in the first stage, a state estimator \textbf{state estimator} $\mathcal{E}^{\text{est}}(o_{:t})$ is trained at the same time to predict the privileged feature extracted by the \textbf{privileged encoder} $\mathcal{E}^{\text{priv}}(o_t, o_t^{\rm priv})$. 
Different from \cite{kumar2021rma}, we initialize the student actor $\pi^{\rm student}$ with the teacher's parameters and continue to train it using PPO in the second stage. This way, the student is driven to find the optimal policy with imperfect estimation instead of faithfully imitating the teacher.

Following the asymmetric actor-critic paradigm, the critic $V(o_t, o^{\text{priv}}_t)$ always receives both observable and privileged information to ensure accurate value estimation.
The policy is trained using PPO \cite{schulman2017proximal} in IsaacLab \cite{mittal2023orbit}.
More implementation and training details can be found in \cref{sec:appx_training}.



\section{Simulation Experiments}
In this section, we conduct comprehensive experiments in simulation to validate the efficacy of our method and design. The domain randomizations used during training, e.g., random perturbations to joint parameters and body mass, are also applied during evaluation.

\begin{figure}[htbp]
    \centering
    \begin{subfigure}[b]{0.329\textwidth}
        \includegraphics[width=\linewidth]{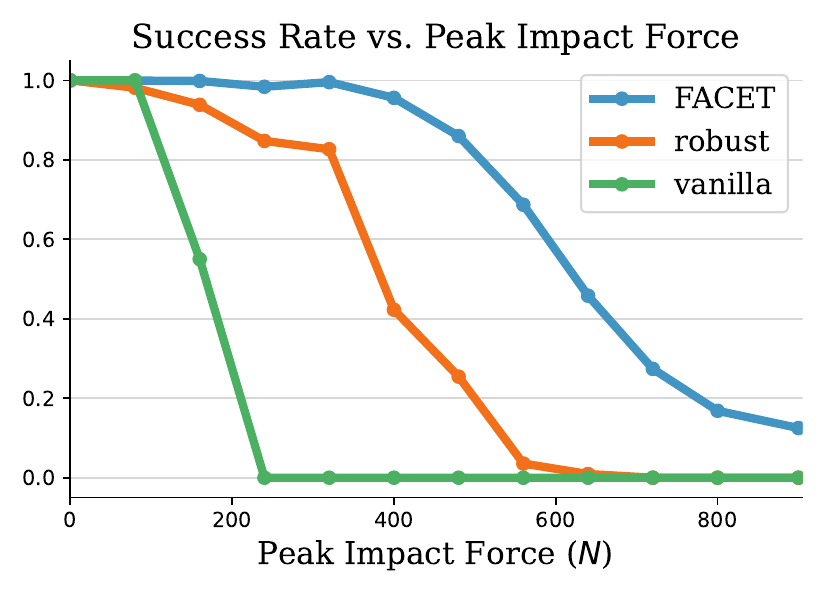}
    \end{subfigure}
    \begin{subfigure}[b]{0.329\textwidth}
        \includegraphics[width=\linewidth]{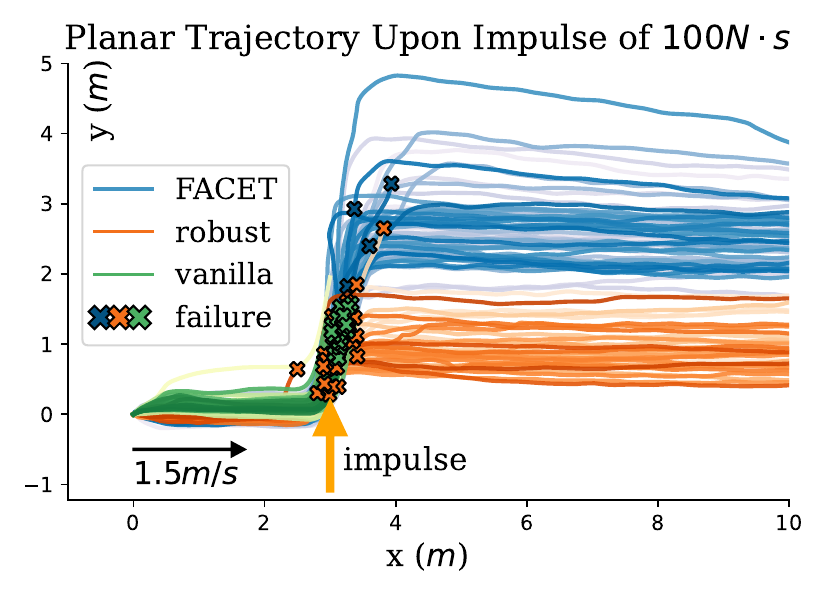}
    \end{subfigure}
    \begin{subfigure}[b]{0.329\textwidth}
        \includegraphics[width=\linewidth]{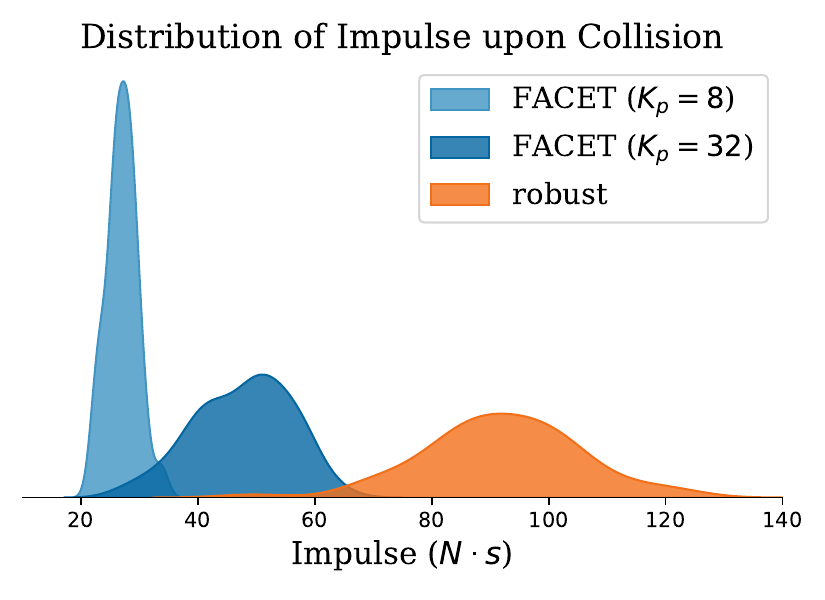}
    \end{subfigure}
    \caption{\textbf{Left}: The success rate under varying levels of lateral impulses during locomotion at 1.5 m/s in the x-direction. \textcolor{blue}{\method{}} demonstrates superior robustness to large impulses compared to baseline policies trained with velocity tracking (\vanilla{}) and with random impulse perturbations (\robust{}). \textbf{Middle}: Planar (xy) trajectories of robot CoM under $400N$ peak force. 64 trajectories are shown for each policy. \textcolor{blue}{\method{}} can compliantly follow the impulse, adapting its velocity to keep balance, unlike the stiffer responses or failures of the baselines. \textbf{Right}: Distribution of collision impulse when the robot walks into a virtual wall. \textcolor{blue}{\method{}} achieves a significantly lower collision impulse, and this can be modulated by adjusting the impedance parameter $K_p$, indicating enhanced safety during physical interaction.}
    \vspace{-5pt}
    \label{fig:simlation}
\end{figure}

\paragraph{Robustness to large perturbations.}
Our method introduces controllable compliance into the policy, allowing it to follow instead of counteract external forces when the forces are too large. To show that, we command the robot to walk forward along the x-axis at $1.5m/s$ and add a horizontal ramp impulse of varying magnitudes.
We compare our method (\textcolor{blue}{\method{}}) with two baselines: (1) \vanilla{}: a locomotion policy trained with velocity tracking reward, and (2) \robust{} trained additionally with random impulse domain randomizations. 
The distribution of perturbations applied during training for \textcolor{blue}{\method{}} and \robust{} are identical (random impulse lasting $0.4\sim0.6s$, peaking $80\sim200N$). Failure is defined as bodies other than feet colliding with the ground.

As shown in \cref{fig:simlation} (Left), \vanilla{} quickly fails as the impulse gets larger. While \robust{} policy shows that random perturbation during training enhances robustness, it is only effective within a medium range. In contrast, \textcolor{blue}{\method{}} can survive even larger impulses. This is depicted in \cref{fig:simlation} (Middle), which shows the trajectory at the peak impact force of $400N$. While \vanilla{} and \robust{} either fail or resist stiffly, \textcolor{blue}{\method{}} adjusts its velocity to move along the y-axis to restore balance. 

\paragraph{Soft collision.} Safety is a critical aspect when the robot operates in cluttered environments, where intense collisions can damage both the subject and the robot itself. Our approach enables more compliant behavior upon collision by using small $K_p$ and $K_d$. We validate this property in simulation by letting the robot walk into a wall (implemented using a soft collision model) at $1.5m/s$ and measure the impulse (change of momentum) within $0.5s$ after collision. As shown in \cref{fig:simlation} (Right), the impulse induced by \textcolor{blue}{\method{}} is significantly smaller than \robust{} and can be controlled by tuning $K_p$.

\begin{figure}[h]
    \vspace{-5pt}
    \centering
    \begin{subfigure}[b]{0.58\textwidth}
        \adjustbox{valign=m}{\includegraphics[width=\linewidth]{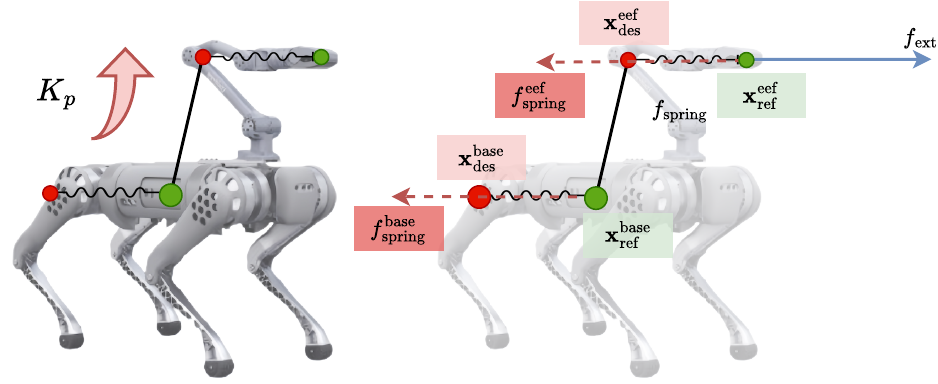}}
    \end{subfigure}
    \hfill
    \begin{subfigure}[b]{0.4\textwidth}
        \adjustbox{valign=m}{\includegraphics[width=\linewidth]{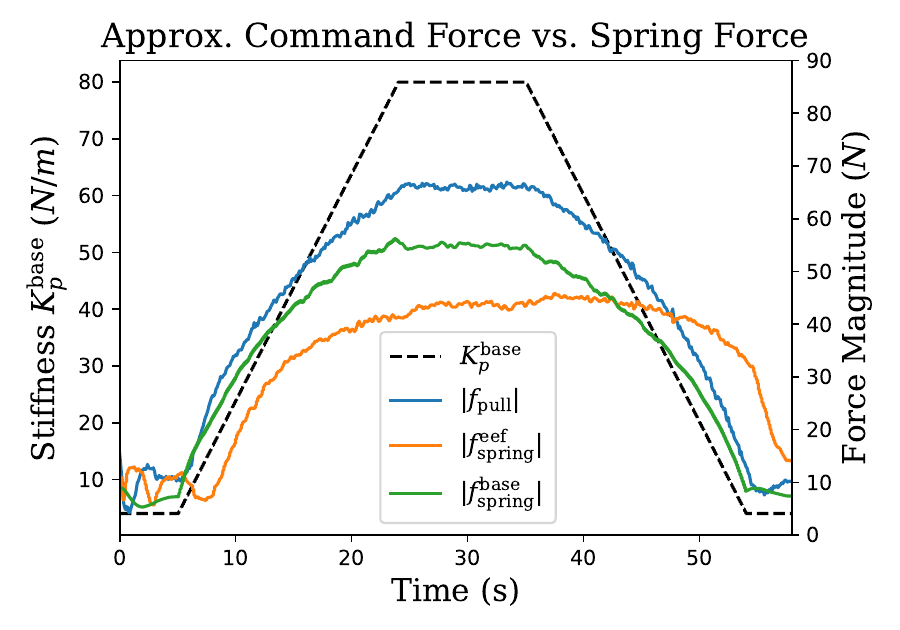}}
    \end{subfigure}
    \caption{Extension to legged loco-manipulator with two bodies of interest. With a fixed $K_p^{\rm eef}=90$, we increase and then decrease $K^{\rm base}_p$ to examine the relationship between the virtual spring forces and the actual pulling forces produced $f_{\rm pull}$. Similar to \cite{portela2024learning}, $f_{\rm pull}$ is approximated by a force applied at the end-effector, which counteracts the robot so that it has near-zero velocity.}
    \vspace{-5pt}
    \label{fig:loco-manip}
\end{figure}

\paragraph{Extension to multiple bodies: legged loco-manipulator.}
On more complex robots, we may be interested in more than the CoM of the trunk body. Our formulation can readily extend to such cases. For example, for a legged loco-manipulator (Unitree B1 quadruped + Z1 Arm), we can define reference models for both the base link and the end-effector: \begin{align}
    m^{\rm base}\refacc^{\rm base} = &  f^{\rm base}_{\rm spring} + f^{\rm base}_{\rm ext} - a\cdot f^{\rm eef}_{\rm spring}, \\
    m^{\rm eef}\refacc^{\rm eef} = & f^{\rm eef}_{\rm spring} + f^{\rm eef}_{\rm ext},
\end{align}
where $a\in[0,1]$ controls whether the force applied on the end effector would transmit to the base. The reward functions are defined analogously to \cref{eq:reward} for end-effector and base, respectively. Setting $a=1$ and $\refpos^{\rm base}=\qpos^{\rm base}$ gives the fully compliant mode. This convenient property allows us to perform kinesthetic teaching by pulling the end-effector to move the robot as shown in \cref{fig:head} (c). On the other hand, we can whole-body large force pulling by specifying high values for both $K^{\rm base}_p$ and $K^{\rm eef}_p$. This is illustrated in \cref{fig:loco-manip}, where we vary the value of $K_p^{\rm base}$ to gradually increase and then decrease the pulling force. The pulling force produced roughly matches the force specified in $f_{\rm spring}^{\rm eef}$ and $f_{\rm spring}^{\rm base}$, showing the desirable property of fine-grained force control.


\section{Ablation Study}

We evaluate two key questions: \textbf{Q1:} Must we use integrated position and velocity trajectories rather than raw acceleration as tracking targets? \textbf{Q2:} Does temporal smoothing of the reference trajectories yield more stable performance compared to open-loop or closed-loop targets? To answer these, we compare four \emph{teacher}s—\texttt{smoothed}, \texttt{open-loop}, \texttt{closed-loop}, and \texttt{acceleration} targets—and three \emph{student} policies derived from the \texttt{smoothed} and \texttt{open-loop} teachers, plus a \texttt{concurrent} baseline. Performance is measured by tracking errors 
and success rate. See \cref{appendix:ablation_metrics} for detailed definition.

\begin{table}[h]
  \centering
  \vspace{-10pt}
  \small
  \caption{Success Rate and Normalized Tracking Errors}
  \label{tab:ablation}
  \begin{tabular}{lcccc}
    \toprule
    \multirow{2}{*}{Method}   & \multicolumn{3}{c}{Normalized Errors $\downarrow$} & \multirow{2}{*}{Success $\uparrow$} \\
                              & pos & vel & acc & \\
    \midrule
    Smoothed Teacher   & 1.00  & 1.00  & 1.00  & 0.98 \\
    Open-loop Teacher   & \textcolor{red_best}{0.72}  & \textcolor{red_best}{0.92}  & \textcolor{red_best}{0.97}  & 0.97 \\
    Closed-loop Teacher & 17.71 & 2.04  & 1.14  & 0.96 \\
    Acceleration Teacher  & 70.51 & 6.64  & 1.65  & 0.97 \\
    \midrule
    Smoothed Student   & \textcolor{red_best}{5.18}  & \textcolor{red_best}{1.85}  & 1.05  & 0.97 \\
    Open-loop Student   & 10.31 & 2.07  & \textcolor{red_best}{1.04}  & 0.98 \\
    Concurrent                    & 26.23 & 3.53  & 1.19  & 0.92 \\
    \bottomrule
  \end{tabular}
  \vspace{-10pt}
\end{table}

For the \emph{teacher} policies, the large position and velocity errors of \texttt{Open-loop Teacher} and \texttt{Acceleration Teacher} indicate they fail to make meaningful progress throughout, while \texttt{Closed-loop Teacher} achieves a similar result as \texttt{Smoothed Teacher}, indicating that integrated position/velocity trajectories are essential for stable reference tracking. The \texttt{Open-loop Teacher} achieves even slightly better performance than the \texttt{Smoothed Teacher}, potentially because the open-loop reference trajectory is smoother and less noisy. However, it does not compensate for the gap between the robot dynamics and the reference model and is therefore too perfect for the student to learn from. This is evidenced by the higher errors by \texttt{Open-loop Student} compared to \texttt{Smoothed Student} after the second stage. 

Acceleration errors remain comparatively large and similar across all variants, confirming that raw \texttt{Acceleration} is not suited to be used as a reward signal. 
The smoothed targets, among others, allow more imperfection and lead to a smaller imitation gap while still adhering to the reference model.

Meanwhile, the one-stage explicit estimation baseline \texttt{Concurrent} \cite{ji2022concurrent}, which directly estimates the external forces and reference targets, also leads to much worse performance. Estimating such information in the input space is challenging by itself, and inaccurate estimation would in turn harm training.

\section{Real-World Results}


In this section, we deploy the student policy to a real Unitree Go2 quadruped for further validation of the force-adaptive capabilities offered by our approach.

\paragraph{Compliant following.}

To demonstrate our policy is compliant and can adapt to very small forces, we set impedance gains $K_p=0, K_d=4$, and apply small external forces to guide the robot's motion. 
As shown in \cref{fig:head} (a), the robot follows effortlessly under gentle pulling with a string attached to its body, and can start or stop moving with the push of a fingertip. 
Despite the absence of explicit velocity commands, the robot smoothly follows the direction of the applied force. This behavior is difficult to achieve with velocity-tracking policies or the built-in controller, which resist deviations from velocity command. 


\paragraph{Large force pulling.}

\begin{wrapfigure}{r}{0.5 \textwidth}
\vspace{-10pt}
    \includegraphics[width=\linewidth]{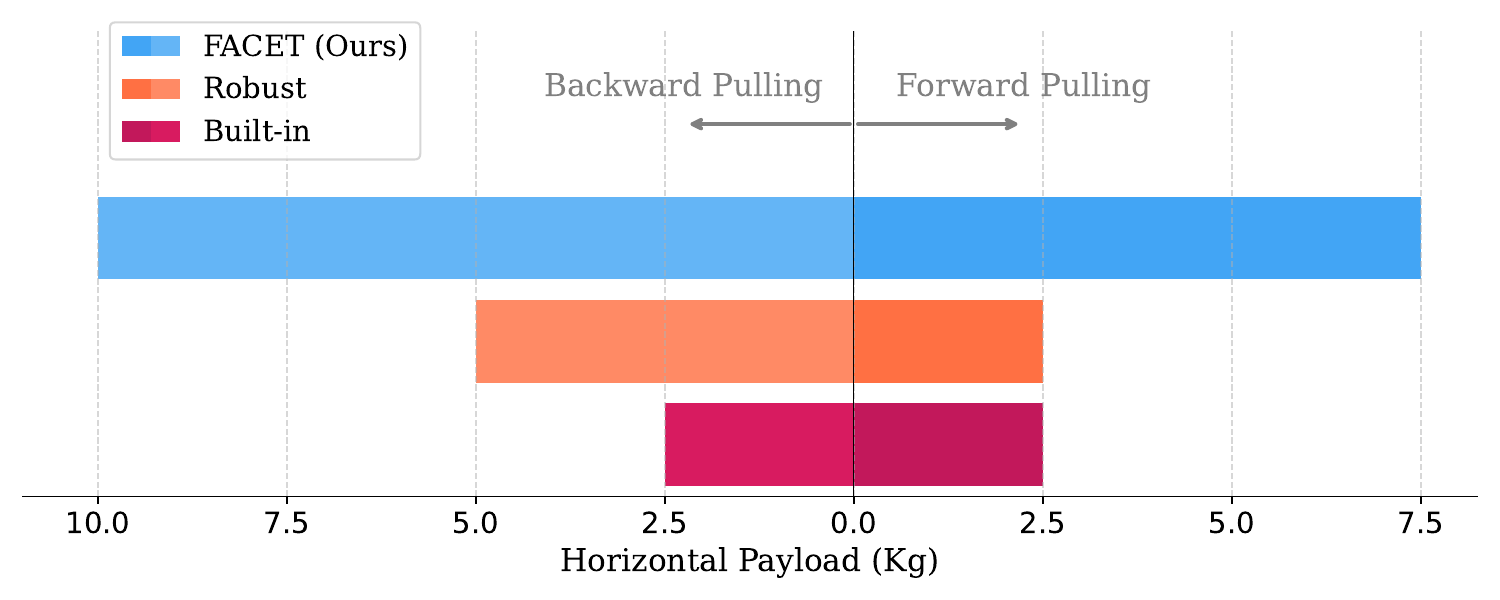}
    \caption{\textbf{Large Force Pulling}. Our controller achieves a significantly wider pulling range than the baselines and demonstrates more robust whole-body coordination behaviors.}
    \label{fig:pulling-barplot}
\vspace{-10pt}
\end{wrapfigure}

To evaluate the force-generating capability and robustness of our controller, we set the impedance gains $K_p$ to a large value. In this setup, the quadruped robot is tethered to a payload via a pulley in gym and tasked with pulling forward/backward. We compare \textcolor{blue}{\method{}} with two baselines: (1) a velocity-tracking policy (\robust), (2) the built-in controller (\builtin).
A video demonstration of the pulling task is provided in the supplementary material.

The built-in controller successfully pulls up to 2.5kg, while \robust{} fails beyond 5 kg, frequently collapsing or spinning in place due to instability. In contrast, our \textcolor{blue}{\method{}} policy reliably pulls up to 10kg without failure, maintaining stable contact and body orientation.

\section{Conclusion}
\label{sec:conclusion}
In this paper, we introduced FACET, a framework for training force-aware controllers for legged robots, and validated its efficacy through experiments. It addresses the issue found in previous learning-based controllers, where responses to forces are often unspecified or cannot be modulated, which we believe to be essential for robots to safely operate in complex environments with human presence. 


\section{Limitations}

Our method has two main limitations: (1) the gap between the robot dynamics and the reference model is not fully addressed, especially when applied in the multi-body setting; and (2) the sim-to-real gap caused by the lack of reliable force sensing and acceleration measurement. Regarding (1), the mechanical constraints between multiple bodies are currently not considered. For example, the reference state targets of the end-effector of a loco-manipulator's end-effector can be infeasible to reach. Regarding (2), better state estimation methods could be incorporated to make full use of the sensors available on a robot, e.g., accelerometer and foot force sensors. Meanwhile, we used oscillators with predefined frequencies for gait control. While providing a clear reward signal for learning stable and nice-looking gaits, it may limit the robot's ability to adapt its foothold based on the target acceleration and velocity. We will move on to resolve these issues in future work.

\acknowledgments{We sincerely thank Chaoyi Pan, Chenhao Lu, Guanqi He, Guanya Shi, Sicheng He, and Zhengmao He (alphabetical order) for their valuable insights and suggestions that made this work more complete.\\
This work is supported by the National Natural Science Foundation of China (62176135), the National Key R\&D Program of China (2022ZD0161700), Shanghai Qi Zhi Institute Innovation Program SQZ202306 and the Tsinghua University Dushi Program, the grant of National Natural Science Foundation of China (NSFC) 12201341.
}


\clearpage
\bibliography{ref}  

\clearpage
\section{Appendix}

\subsection{Implementation and Training Details}

\label{sec:appx_training}

\subsubsection{Training Setup}
The training environment and the task setup are implemented based on Isaac Lab \cite{mittal2023orbit}. We use PPO \cite{schulman2017proximal} to train the policies with 4096 parallel environments. Following the setup shown in \cref{fig:teacher-student}, the critic $V$ and encoder $\mathcal{E}$ are Multi-Layer Perceptron (MLP) with $(512, 256, 256)$ and $256$ hidden units, respectively. The estimator $\hat{\mathcal{E}}$ is a GRU \cite{chung2014empirical} with hidden dimension 256, while the policy $\pi$ is parameterized by an MLP with $(256, 256, 256)$ hidden units outputting the mean of a diagonal Gaussian distribution with state-independent variance. Empirical normalization with running statistics is applied to the network inputs to improve optimization. The two-stage training runs for 200 million environment steps (first stage) and 50 million steps (second stage), requiring approximately 1 hour on a gaming laptop with \text{Intel Core i9-13980HX CPU} and \text{NVIDIA RTX 4080 GPU}. More details, such as hyper-parameters, can be found in the codebase.

\subsubsection{Observation Space Details}
\label{appendix:observation-details}

We adopt a teacher-student observation structure where the student has access to a limited set of observations suitable for real-world deployment, while the teacher is granted access to privileged information during simulation training. All Cartesian-space observations, including position, and velocity, and wrench, are converted to the base's frame.

\begin{table}[h]
\centering
\caption{List of Observations}
\small
\label{tab:observations}
\begin{tabular}{@{}llll@{}}
\toprule
\textbf{Observation} & Notation & \textbf{Available To} & \textbf{Description} \\
\midrule
Command & $\mathbf{u}$ & Student \& Teacher & Includes $\qpos_{des}$, $K_p$, $K_d$, and virtual mass $\mathbf{m}$ \\
Joint Positions & $\mathbf{q}$ & Student \& Teacher & Raw joint angles \\
Joint Velocities & $\dot{\mathbf{q}}$ & Student \& Teacher & Angular velocities of joints \\
Base Angular Velocity & $\omega$ & Student \& Teacher & Angular velocity of the base (from IMU) \\
Projected Gravity Vector & $g$ & Student \& Teacher & Gravity vector projected into base frame \\
\midrule
Base Linear Velocity & $\mathbf{x}$ & Teacher Only & Linear velocity of the robot’s base \\
Feet Contact States & $\mathbf{c}$ & Teacher Only & Binary indicators of foot-ground contact \\
External Forces/Torques & $f_{\rm ext}$ & Teacher Only & Forces and torques applied to the base \\
Joint torques & $\boldsymbol{\tau}$ & Teacher Only & Torques Commanded/applied at each joint \\
Reference Targets & $\{\refpos,\refvel\}$ & Teacher Only & Reference targets described in \cref{sec:temporal_smoothing}\\
\bottomrule
\end{tabular}
\end{table}

\subsubsection{Rewards Functions}

The main task reward consists of position and velocity tracking of xy translation and yaw rotation with the form and tracking targets described in \cref{eq:reward}. The loco-manipulation experiment additionally includes position and tracking rewards defined for the end-effector. In practice, we found tracking rewards of the form $\exp(-||\qvel-\refvel||^2_2)$ may saturate when the error is large during the early stage. A quadratic cost term $-||\qvel-\refvel||^2_2$ is added to resolve this issue. A series of regularization rewards is used to ensure natural behaviors. A full list of rewards for the Unitree Go2 quadruped task is summarized in \cref{tab:rewards}.

\begin{table}[h!]
    \centering
    \small
    \begin{tabular}{lll}
        \toprule
        \multicolumn{3}{c}{\textbf{Impedance Tracking Rewards}} \\
        \midrule
        Reward Name & Weight & Function \\
        \midrule
        impedance\_yaw\_pos & 0.75 & $\exp(-\|x_{\rm ref}^{\rm yaw} - x_{\rm sim}^{\rm yaw}\|^2 / 0.25)$ \\
        impedance\_yaw\_vel & 1.0 & $\exp(-\|\dot{x}_{\rm ref}^{\rm yaw} - \dot{x}_{\rm sim}^{\rm yaw}\|^2 / 0.25) - 0.5 \|\dot{x}_{\rm ref}^{\rm yaw} - \dot{x}_{\rm sim}^{\rm yaw}\|^2$ \\
        impedance\_pos & 0.75 & $\exp(-\|{\mathbf{x}}_{\rm ref}^{xy} - {\mathbf{x}}_{\rm sim}^{xy}\|^2 / 0.25)$ \\
        impedance\_vel & 1.5 & $\exp(-\|{\mathbf{v}}_{\rm ref}^{xy} - {\mathbf{v}}_{\rm sim}^{xy}\|^2 / 0.25) - 0.5 \|{\mathbf{v}}_{\rm ref}^{xy} - {\mathbf{v}}_{\rm sim}^{xy}\|^2$ \\
        \midrule
        \multicolumn{3}{c}{\textbf{Regularization Rewards}} \\
        \midrule
        angvel\_xy\_l2 & 0.02 & $-\|\boldsymbol{\omega}_{\rm sim}^{xy}\|^2$ \\
        linvel\_z\_l2 & 2.0 & $- v_{z, \rm sim}^2$ \\
        base\_height\_l1 & 2.0 & $-|z_{\rm sim} - 0.35|$ \\
        energy\_l1 & 0.0002 & $-\|{\boldsymbol{\tau}} \odot \dot{\mathbf{q}}\|_1$ \\
        joint\_acc\_l2 & $2.5 \times 10^{-7}$ & $-\|\mathbf{\ddot{q}}_{\rm sim}\|^2$ \\
        joint\_torques\_l2 & $2.0 \times 10^{-4}$ & $-\|\boldsymbol{\tau}_{\rm sim}\|^2$ \\
        survival & 1.0 & $1$ \\
        action\_rate\_l2 & 0.01 & $-\|\dot{\mathbf{a}}\|^2$ \\
        feet\_air\_time & 0.4 & $\sum_{\rm num\_feet} \max(0, t_{\rm air} - 0.3)$ \\
        undesired\_contact & 2.0 & $-\sum_{\rm undesired\_bodies} I(\text{contact})$ \\
        \bottomrule
    \end{tabular}
    \label{tab:rewards}
\end{table}

\subsubsection{Command and Force Sampling}
\label{appendix:sampling_details}

\paragraph{Command.}Following \cref{eq:ref_model}, we sample impedance parameters $\mathbf{u}=(\qpos_{des}, K_p, K_d, \mathbf{m})$ and external forces $\mathbf{f}_\text{ext}$ that induce force-adaptive behavior in forceful interactions. To ensure natural motion and avoid oscillation around equilibrium, we always use the critical damping $K_d=2\sqrt{K_p}$. The virtual mass is sampled from $\{1, 2, 4\}kg$. For the quadruped task, the setpoint $\qpos_{\rm des}$ is periodically sampled and updated in one of the following ways:
\begin{enumerate}[leftmargin=*]
    \item \textbf{Fixed Target Position}: A new setpoint is sampled around the current base position $\qpos_{\rm sim}$ in the world frame and stays fixed for a few seconds before the next one is sampled.
    \item \textbf{Fixed Target Velocity}: A target velocity $\qvel^{\rm target}$ is sampled and the setpoint is updated at each step as $\qpos_{\rm des}=\frac{K_d}{K_p}\qvel^{\rm target} + \qpos_{\rm sim}$ so that the velocity converges to $\qvel^{\rm target}$.
    \item \textbf{Fully Compliant}: The setpoint moves with $\qpos_{\rm sim}$, i.e., we set $\qpos_{\rm ref} := \qpos_{\rm sim}$ at each step. The virtual spring force $f_{\rm ext}$ stays zero so that the robot moves fully compliant to external forces.
\end{enumerate}

For the loco-manipulation task, we sample the end effector setpoint $\qpos^{(1)}_{\rm des}$ in the body frame of the quadruped base. To ensure the setpoint is feasible to reach (subject to some error threshold), an archive of such poses is obtained in advance using Inverse Kinematics. In this work, the end-effector's desired orientation is always set as forward in the base's frame. 

\paragraph{External forces.}
To simulate forceful interactions with the environment, three types of forces are randomly sampled and applied:
\begin{enumerate}[leftmargin=*]
    \item \textbf{Constant}: A steady force is sampled from a predefined range and applied continuously for a $1\sim4s$ time. For the quadruped task, the maximum force along $x$, $y$ and $z$ axis are $40N$, $40N$ and $15N$, respectively.  
    \item \textbf{Impulse}: A high-magnitude force applied for a short duration $0.4\sim0.6s$. We adopt a ramp impulse that linearly increases to the peak and then decreases to zero. The resulting impulse is therefore $0.5 \cdot \text{duration} \cdot f_{peak}$. For the quadruped task, the peak forces along the $x$, $y$, and $z$ axes are sampled from $80\sim200N$, $80\sim200N$, and $0\sim20N$, respectively.
    \item \textbf{Spring}: A force that behaves as if the point of interest is attached to a highly stiff spring similar to that described in \cite{portela2024learning}.
\end{enumerate}
We also introduce small positional offsets relative to the center of mass so that these forces produce torques as well.

\subsection{Simulation Experiments}
\label{appendix:sim_exp}

\paragraph{Robustness to large perturbations.}We command the robot to walk forward along the x-axis at $\qvel^{\rm target} = 1.5m/s$ by setting the setpoint in front of the robot according to $\qpos_{\rm des}=\frac{K_d}{K_p}\qvel^{\rm target} + \qpos_{\rm sim}$. We add the horizontal ramp impulse at $t=2.0s$, i.e., after the locomotion is stable. The impulse lasts for $0.4\sim0.6s$ and has varying magnitudes. For each policy, we evaluate the policy in 64 parallel environments with randomizations. Failure is defined as bodies other than feet colliding with the ground. The \textcolor{temp_red}{robust} baseline is trained with the same random impulse described above as our method. 

\paragraph{Soft collision.}We implement the virtual wall with a soft collision model, i.e., the robot will experience an external force when colliding into the wall:
\begin{equation}
    f_{\rm wall} = \begin{cases}
        - K_p^{\rm wall} d_{\rm penetration} - K_d^{\rm wall} \dot d_{\rm penetration}, d_{\rm penatration} > 0\\
        0, d_{\rm penetration} \le 0
    \end{cases}
\end{equation}
where $d_{\rm penetration}$ is the distance of penetration normal to the wall plane direction, measured at the robot's base. We measure the impulse by integrating the wall contact force $f_{\rm wall}$ within $0.5s$ after collision takes place. 


\subsection{Ablation Study}
\label{appendix:ablation_metrics}

We random sample the impedance parameters $\qpos_{\rm des}, K_p, K_d$ and external force $f_{\rm ext}$ following the strategy described in \cref{appendix:sampling_details}. 

For all policies trained with different tracking targets, we measure tracking error with a unified metric: the position and velocity tracking error is computed with respect to the reference targets $\qpos_{\rm tef}^{t'}, \qvel_{\rm tef}^{t'}$ for $t'\in\{t-8\Delta t, t-16\Delta t,t-32\Delta t,0\}$ and the acceleration error is computed against the closed-loop target $\qacc_{\rm ref}^{t-1}$:
\begin{align}
    \text{pos error}=& \sum_{t'} ||\refpos^{t'}-\qpos||^2_2, \\
    \text{vel error}=& \sum_{t'} ||\refvel^{t'}-\qvel||^2_2, \\
    \text{acc error}=& ||\refacc-\qacc||^2_2.
\end{align}

Such metrics are chosen because we want the policy to resample the reference model as close as possible, while accurately following the \textcolor{temp_red}{open-loop targets} $(\refpos^0,\refvel^0)$ is infeasible and also unnecessary. Similar to reward computation, we use mixtures of targets for the position and velocity errors. As for acceleration, we directly compute the error w.r.t the reference at that step since we care more about its instantaneous behavior.

\subsection{Real-World Experiments}
\label{appendix:real_exp}

In most cases, we do not have access to a well-defined world frame on the hardware. When deployed to a real robot, we can only specify the translation part of the setpoint $\qpos_{\rm des}$ directly in the robot's frame following the \textbf{fixed target velocity} mode or \textbf{fully compliant mode} described in \cref{appendix:sampling_details}. The yaw rotation part of $\qpos_{\rm des}$ can be specified in the rotation frame created when the robot starts up.  

\paragraph{Compliant following.}
The compliant following behavior can be produced with the \textbf{fully compliant mode} with any valid $K_d$ within the training range. A smaller $K_d$ makes it easier to drive the robot.
\paragraph{Large force pulling.}In the large force pulling experiment, we gradually increase $K_p$ and setpoint displacement so that $K_p(\refpos-\qpos)$ increases to $100N$. Videos and further results can be found in the supplementary materials.

\end{document}